# VacciNet: Towards a Smart Framework for Learning the Distribution Chain Optimization of Vaccines for a Pandemic

**Jayeeta Mondal**
Embedded Device & Intelligent Systems group
TCS Research
Kolkata, West Bengal, India

**Jeet Dutta**
Embedded Device & Intelligent Systems group
TCS Research
Kolkata, West Bengal, India

**Hrishav Bakul Barua**
Robotics & Autonomous Systems group
TCS Research
Kolkata, West Bengal, India
hbarua@acm.org

## Abstract

Vaccinations against viruses have always been the need of the hour since long past. However, it is hard to efficiently distribute the vaccines (on time) to all the corners of a country, especially during a pandemic. Considering the vastness of the population, diversified communities, and demands of a smart society, it is an important task to optimize the vaccine distribution strategy in any country/state effectively. Although there is a profusion of data (Big Data) from various vaccine administration sites that can be mined to gain valuable insights about mass vaccination drives, very few attempts has been made towards revolutionizing the traditional mass vaccination campaigns to mitigate the socio-economic crises of pandemic afflicted countries. In this paper, we bridge this gap in studies and experimentation. We collect daily vaccination data which is publicly available and carefully analyze it to generate meaning-full insights and predictions. We put forward a novel framework leveraging Supervised Learning and Reinforcement Learning (RL) which we call *VacciNet*, that is capable of learning to predict the demand of vaccination in a state of a country as well as suggest optimal vaccine allocation in the state for minimum cost of procurement and supply. At the present, our framework is trained and tested with vaccination data of the USA.

## 1 Introduction

Infectious diseases and pandemic outbreaks have impacted the world history by shaping societies, changing war outcomes, influencing socio-economic policies and political standpoints, and overall paving the way for innovation in medicine and technology [1]. A growing consensus among researchers is that, one of the most effective way of controlling pandemic outbreaks is a well-timed and efficiently managed vaccination strategy. In the phase-1 of *Covid-19* vaccination, the doses were made available to the most vulnerable population, based on age and occupation. The United States (USA) Department of Health and Human Services adopted the Operation Warp Speed (OWS) strategy for Covid-19 vaccine distribution in January 2021 [2], which does not consider several important optimization factors. According to the recent updates in the World Health Organization (WHO) Covid-19 dashboard, globally there have been 5 billion vaccinations administered. Currently 2 billion people, which is only 29.9% of the world's total population, are fully vaccinated. The mutations of the *SARS-CoV-2* virus has made development of mutation-resistant vaccines and administration of seasonal vaccinations inevitable in the near future [3]. Till date mass vaccinations for Covid-19 booster doses are getting carried out, while emergence of new strains and viral outbreaks remain a high possibility. Figure 1 shows a typical cold chain vaccine distribution network for a country. For delivering 1 billion Covid-19 vaccines USA spent nearly **2 billion US dollars**, about **22%** of which is associated with Controlled Temperature Chain (CTC) transportation [4]. Frozen essential goods need high maintenance to avoid damage as well. Even in front-line health centres, there is a cost for refrigerated storage of the supplied vaccines. If we continue to conduct uncoordinated vaccine distribution, global GDP loss in the coming time can amount to **US$ 9.2 trillion** [5]. At the present, no automated approach to commercially solve this issue has come up that can help mitigate economic crisis due to mismanaged CTC supply-chain. From our observation of the world-economics during the past two years, it is evident that an automated solution to optimized vaccine or other CTC good supply network can essentially reduce wastage of human and monetary resources by big pharmacies, governments and other companies. Hence, in this paper we attempt to propose a solution with deep learning based approach. We limit our experiments and solutions to vaccine supply chain in this paper, but our framework can be implemented on any CTC essential item supply chain in a country.

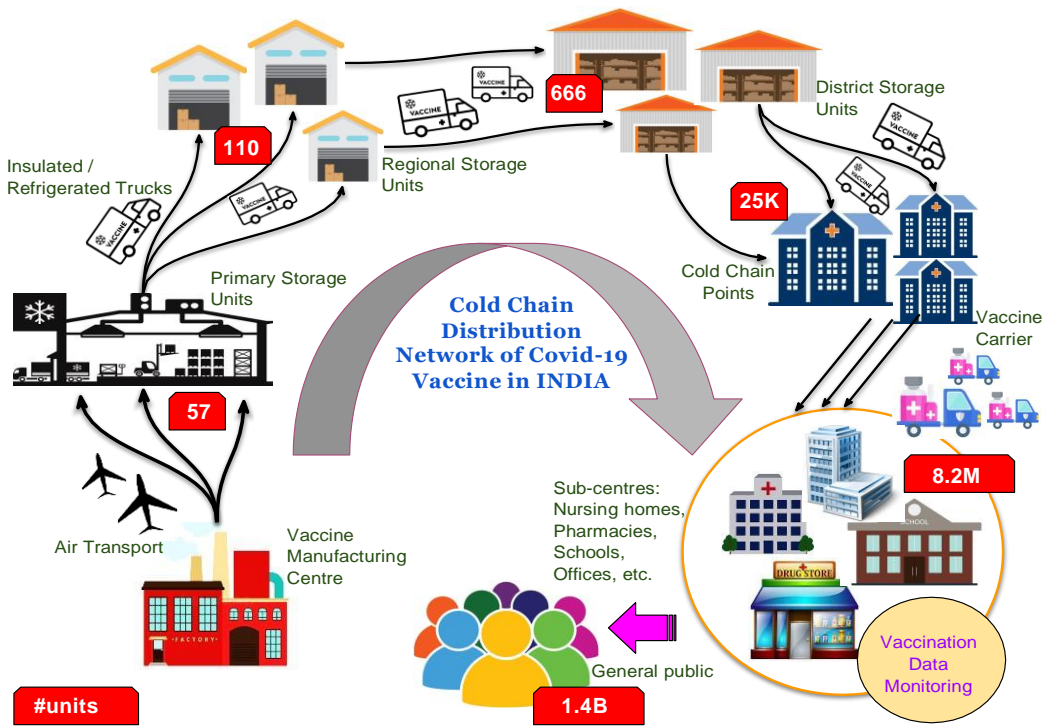

Figure 1: A typical Vaccine Distribution Cold Chain network. Here, the unit counts are from data of India (an example case) [6].

With the aforementioned motivations, our main highlights and contributions in this paper are three-fold –

1. We implement a Recurrent Neural Network (RNN) [7] based vaccine demand prediction model. We use supervised learning mechanism to train a light-weight deep SRU (Simple Recurrent Unit) [8] model to predict the state-wise daily vaccination requirements based on the collected vaccination drive data of different states in the US,

2. We perform feature engineering on the collected vaccination data to help train our SRU predictor model with an added attention block to increase the model convergence speed while training.

3. We implement a Reinforcement Learning (RL) based Deep Q-network [9] to optimize the state-wise vaccine allocation as per the state-wise demands and propose a training algorithm for the same.

The rest of the paper is arranged as follows – in Section 2 we discuss the existing research in this area and explain how our work and contribution is different from the earlier works. We introduce our working methodology and proposed framework in Section 3 and finally provide the evaluation results of our framework in Section 4. Finally, we discuss the comparison of our work with the earlier works, the limitations as well as the future prospects of our work in Section 5.

## 2 Related Works

Vaccine distribution network primarily depends on vaccine coverage, release time, and deployment methods [10]. Some Food and Drug Administration (FDA) approved vaccines need expensive CTC handling for thermostability and hence need accurately optimized procurement, distribution, and storage management strategies to help curb global pandemic and economic crisis [11] [12] [13]. Babus *et al.*[14] examined occupation based infection risks and age based fatality risks of general population to model an optimal vaccine allocation strategy. Matrajit *et al.*[15] evaluated selective objective functions to estimate vaccination coverage and effectiveness.

A concatenation of Reinforcement Learning (RL) and Contextual Bandits sub-models in feed-forward has been proposed in [16] for phase-1 Covid-19 vaccination distribution. The authors have used death-rate, recovery-rate, hospital facilities, etc. of a state as input attributes to predict which part of the population is at a higher risk of pandemic fatalities and therefore needs priority vaccination. Although several optimization policies have been adopted for prioritizing vaccine allocation at an early stage of a pandemic (phase-1), there exists a research gap in developing economically optimal vaccine distribution strategies during mass vaccination (phase-2). This paper attempts to bridge the said gap and to the best of our knowledge *VacciNet* framework is the first attempt at this.

The major problems associated with mass vaccination drives are dense, highly mismanaged cold chain distribution networks that can incur unnecessarily large expenditures by governments and health institutions over a very small span of time [4] [5]. In a country that is suffering or that has suffered from the economic damages due to a pandemic, an efficient vaccine or essential item supply chain can help save resources. Since frozen products like vaccines can have huge supply chain costs, effective methods of distribution strategies are extremely critical. We address this problem with our novel deep-learning based methodology detailed in the section below.

## 3 Our System: Overview and Design

In this section of the paper we explain the workflow of our method, detail the data-flow in proposed framework, elaborate the architecture design and mathematical working of the same, and finally provide the implemented learning algorithm.

### 3.1 Proposed Framework

We provide a detailed discussion of the methodology adopted, Deep Learning (DL) models implemented, and their architecture components in this section.

#### 3.1.1 Design Components and Operation

Figure 2 shows the basic block diagram of our proposed framework. We include two stages in our method – 1) Supervised Learning of a RNN-based *predictor* model and 2) Reinforcement Learning. We collect and pre-process the vaccination data of each state in US [1] to create the time-series input feature space for training RNN-based SRU model. We further detail the data preparation in Section 4.1. This SRU model trained and tested on real vaccination data, is capable of predicting the daily vaccination demands of each state in the US. We call this model the *predictor* model. The aim of the *predictor* is to find the daily vaccination requirement of a state based on the state's population and total people vaccinated till a stipulated date. The next step of our framework includes training a deep Q-network (*RL-agent*) with Reinforcement learning such that it can generate the most optimal vaccine allocation per state with minimum cost of vaccine supply-chain. In a reinforcement learning system, the *environment* is the entity with which an *RL-agent* interacts to gain a set of valid observations, also called the *state-space* of the RL algorithm. The interactions of the *RL-agent* makes it take certain actions in the *environment* as permissible in the defined *action-space* of RL algorithm. These actions get better with each iteration of training the *RL-agent*, such that the agent learns reach the best probable state in the over-all *state-space*. We have listed the over-all process in Algorithm 1. However, the last line of the algorithm can be explained with the reinforcement learning algorithm, which is explained in Section 3.2 and Algorithm 2.

---

[1]https://ourworldindata.org/us-states-vaccinations

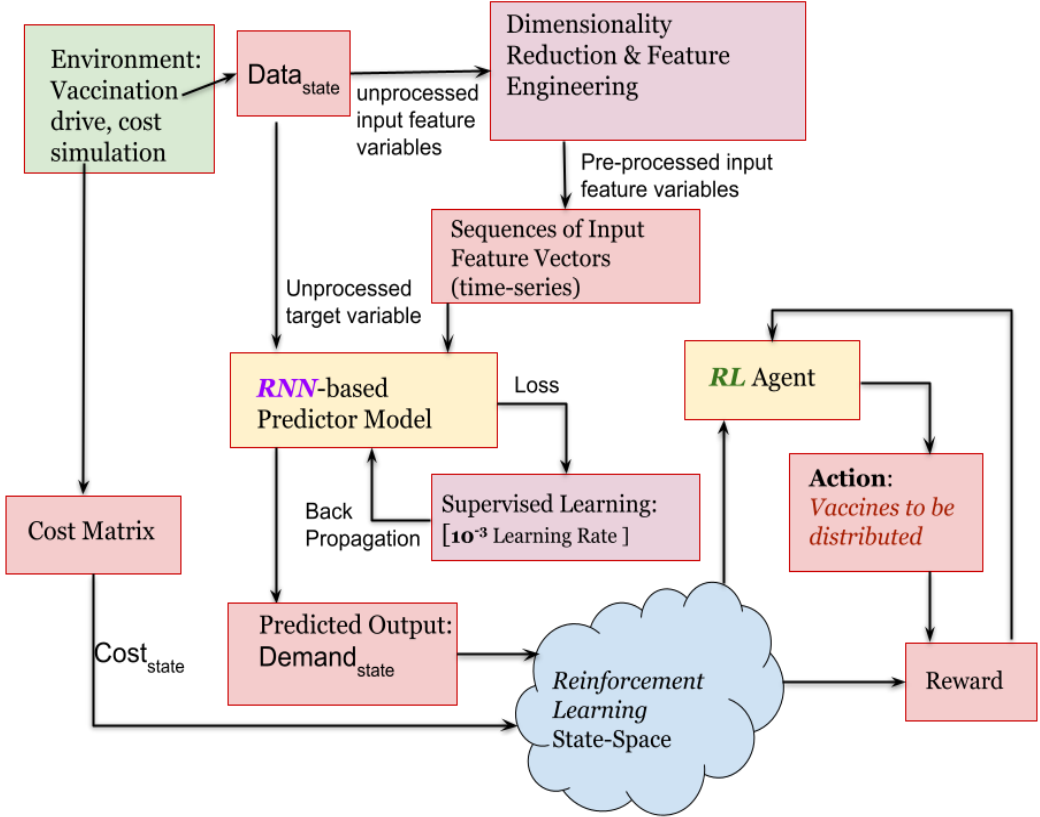

Figure 2: Overview of VacciNet.

In our system we defined the *environment* as – 1. the vaccine demands in each state of a country, and 2. the vaccine supply-chain costs associated to meet the demands. The vaccine demands ($Demand^{state}$) are the output from the trained *predictor*. We simulated the *cost matrix* data considering several real-world parameters that impacts the cost of temperature controlled supply-chains. This data simulation is further explained in Section 4.1.2 in the later part of the paper. Algorithm 1 helps elaborate the workflow of the framework while the complete mathematical foundations of our proposed method is discussed in Section 3.2.

*3.1.2* **Choice of the *Predictor* Model and the *RL-Agent***

As explained previously, we attempt to create two models – a *predictor* model to infer the daily vaccination demands of a state, and a *RL-Agent* which can optimize the vaccine distribution chain. The vaccination data is in the form of a time-series, where one data-point is dependent on the previous set of data points. We use a deep Simple Recurrent Unit (SRU) [8] as the *predictor* model which is a recently introduced variant of RNN that has advantages over Quasi RNNs (QRNNs) [17] and Kernel Neural Networks [18] for their "light recurrence" connections with relatively less parameters [8]. The "highway connections" in SRUs give them an edge over the popular Long Short Term Memories (LSTMs) [19] and Gated Recurrent Units (GRUs) [20] with improved training and inference time. Hence, we select SRU units to construct our RNN-based predictor model that allowed us to reach effective results with minimal effort in training it. We utilize an attention mechanism [21] in SRU *predictor* training that allows us to converge the model **2.9 times faster** than standard SRU model. Although a multitude of techniques have been adopted in the earlier works for supply chain optimization methodologies [22], we propose to use deep Q-networks (DQNs) [9] to optimize the distribution chain using Q-learning. Q-learning has picked up popularity in most recent areas of research due to the sustainability, reliability and high accuracy of the learning method. Q-learning models are easy to implement, train and deploy alongside give high performance compared with other techniques.

*3.1.3* **Model Architecture Details**

Figure 3 shows the detailed data-flow diagram of our proposed framework. As illustrated in Figure 3, our deep *predictor* model consists of two SRU [8] layers (*SRU Layer 1* and *SRU layer 2*) with 10 units each, and a single unit dense layer. The inputs to the SRU units are pre-processed feature attributes, *total_population* (state population), *people_partially_vaccinated* and *people_fully_vaccinated*. We also include an *attention* block [21] in getting the final output *y* which is the predicted daily state-wise vaccine demand (previously explained *Demand_state*). The *attention* block generates *context vector* from the hidden layer output *feature vector*. This helps the model behaviour (inter-mediate feature representation) to focus on important parts of the *input*, before it reaches the *fully connected layer*. For the DQN *RL-agent* we train a deep network of 4 dense layers in RL based Q-learning algorithm. The details of the working of a single SRU-unit (as shown in Figure 3) and the mathematical foundations of our RL algorithm are detailed in Section 3.2. The details of data pre-processing and preparation for the SRU *predictor* and the DQN *RL-agent* are discussed in Section 4.1.

---

**Algorithm 1:** Vaccine distribution chain optimization

**Data:** state-wise vaccination data, state-wise cost matrix data
**Result:** state-wise vaccine allocation
*Cost_state* ←{ state-wise cost matrix } ;
*Data_state* ←{ state-wise vaccination data };
*Allocation_state* ←{ $\phi$ } ;
*Demand_state* ←{ $\phi$ } ;
**for** *index_state, input_features in Data_state* **do**
| *Demand_state[index_state]* ← *SRUPredictor(input_features)*;
**end**
*statespace* ←{*Demand_state, Cost_state* } ;
*Allocation_state* ←*RLagent(statespace)*;

---

### 3.2 Mathematical Formulations and Training Algorithm

The vaccination data collected from vaccination drives in the US consists of data on number of people partially vaccinated, fully vaccinated, and the state-wise doses administered in a day. The state-wise vaccination done (*daily_vaccinations*) is our target variable, and hence remains an unprocessed continuous numerical data. Whereas, we pre-process the other two variables along with state population data to generate final input feature space of SRU *predictor* (details in Section 4.1). The input data (as shown in Figure 3) represents a sequence of features, *x*, over a set of contiguous dates per state. We represent the input as $x = \{x_0, x_1, x_2, ..., x_T\}$, where $x_t$ is a set of feature for the timestep/date *t*. *T* represents the upper bound to the sequence. We can then formulate SRUs as shown in Equations 1, 2, 3 and 4.

The output of the SRU layer is given by $h_t$ in Equation 4 and $f_t$ represents a light recurrence computation in Equation 1, which is highly parallelizable due to the point-wise multiplication($\odot$ ) of $V_f$ and $c_{t-1}$. In Equation 3, $r_t$ represents a highway connection and $c_t$ in Equation 2 represents the state of the SRU cell. Coupled with an *attention* mechanism, the model is trained using a *mean-squared error* loss against the target feature vector *daily_vaccinations*, *y*, of an area using supervised learning method.

$$f_t = \sigma(W_f x_t + V_f \odot c_{t-1} + b_f) \quad (1)$$

$$c_t = f_t \odot c_{t-1} + (1 - f_t) \odot W x_t \quad (2)$$

$$r_t = \sigma(W_r x_t + V_r \odot c_{t-1} + b_r) \quad (3)$$

$$h_t = r_t \odot c_t + (1 - r_t) \odot x_t \quad (4)$$

For the next step of our proposed framework, the RL phase, we take the output (*Demand_state*) of *predictor* as one constituent of *state-space*, whereas for the cost-matrix in the *state-space* we utilize Equation 10 in Section 4.1.2. This data is generated for daily vaccinations in each state of the country.

$$s_{state} = \langle Cost_{state}, Demand_{state} \rangle \quad (5)$$

$$Q^*(s, a) = E[r + \gamma \max_{a'} Q^*(s', a')] \quad (6)$$

Given a state space $s_{state}$, the agent follows a policy $\pi$ with which it produces an action *a*. From this action, it receives a reward given by $Q_\pi(s, a)$. The model tries to maximize its reward by optimizing its *action space* thereafter represented by $Q^*(s, a)$, which is defined in Equation 6. The variables *r* and *γ* are the immediate reward and the discount factor respectively, whereas *s′* and *a′* represents the subsequent permissible state and action spaces of the *RL-agent* model. When the Q-function is represented by a neural network, it can cause instabilities and often diverge from target solution. This happens due to the instabilities in correlation coefficients between Q-value and target values and minor alterations in the Q-values causing drastic changes in policy. Hence, Mnih *et al.*[9] introduced experience replay and iterative updating in Q-learning. In this concept, the *RL-agent* has a collection of experiences called memory. We define experience $e_t$ at time-step $t$ as ($s_t$, $a_t$, $r_t$, $s_{t+1}$) and memory $D_t$ at time step $t$ as $e_1, e_2, \ldots, e_t$. Here, *s*, *a*, and *r* refers to the state-space, action, and reward at time-step *t* respectively. Each *state-space* is formulated by Equation 5. We use a neural network with weight vector $\vartheta$ as Q-network in our Q-learning algorithm. At each iteration *k* of our iterative weight update method, we train the Q-network to reduce the root mean square log error in the Bellman Equation 6. We use root mean square log error (RMSLE) as loss function to update weights in Q-network as given in Equation 7. RMSLE is more robust to outliers in the data. Since the demand in a state or costs of vaccine allocation can be largely different on some days, the real data of this sort is expected to have multiple outliers. Hence, we carefully choose the loss function.

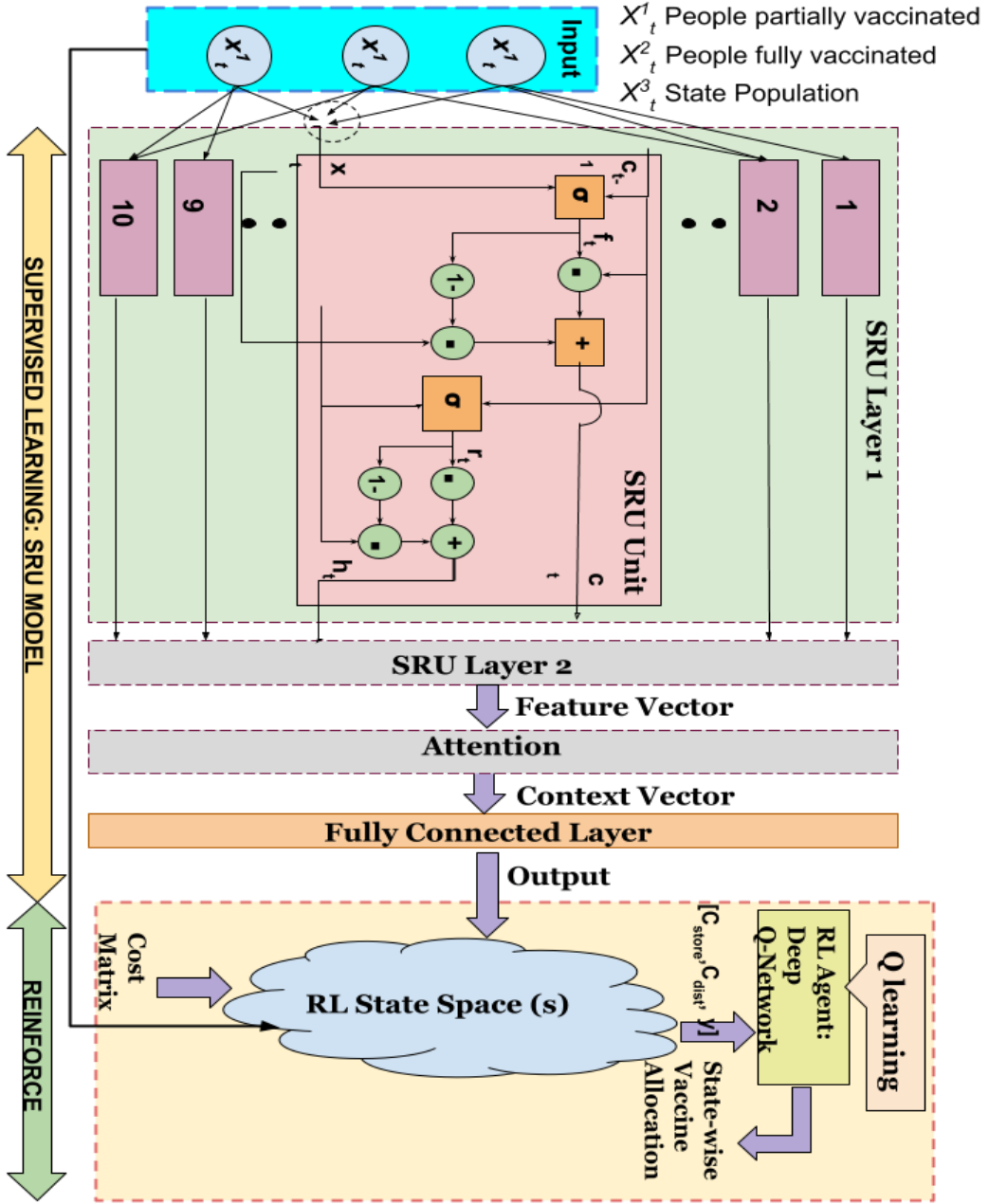

Figure 3: The detailed data-flow diagram of *Vaccinet*. The input vector represents the $Data_{state}$ variable that consists of 3 components – people partially vaccinated ($X^1_t$), people fully vaccinated ($X^2_t$) and state population ($X^3_t$). The output vector represents the $Demand_{state}$ variable. Cost matrix is the $Cost_{state}$.

$$L_k(\vartheta_k) = E_{s,a,r}\left[\sqrt{\frac{1}{K} * \sum_{k=1}^{K}(\log(E_s[y|s,a]) - \log(Q(s,a,\vartheta_k)))^2}\right] \quad (7)$$

In Equation 7, $\vartheta_k$, and $Q$ are the weight vector, and Q-value at $k^{th}$ iteration respectively. The aim of the Deep Q-Learning is to find an optimal policy $\pi$ such that the total reward is maximized, this can be represented by the expected discount return ($G_t$) function in Equation 8. Our reward function is multi-objective in nature, and the RL algorithm tries to reach a trade-off between the cost for vaccine distribution in a state $Cost_{state}$ and the vaccine allocation in that state $Demand_{state}$ as per the demand. Hence in the reward we apply a piece-wise linear [23] negative function [24] over cost data and positive function for vaccine allocation, that is, we start giving negative or positive rewards for cost exceeding predefined threshold and vaccine allocation lower than demand for the state respectively. Hence, we train the deep Q-network to maximize the total reward given by Equation 8.

$$G_t = \sum_{k=0}^{\infty} \gamma^k r_{t+k+1} \quad (8)$$

As shown in algorithm 2, vectors $Cost_{state}$ and $Demand_{state}$ constitute the *state-space s* of RL. The DQN *RL-agent* is repeatedly trained with data of each state, which we consider to total at $S$ states. The function $\phi$ represents the pre-processing performed in the state vector at a particular time-step $t$. We train our *RL-agent* with mini-batches of size 16. For the data of each state in the US, we initiate the training with random actions by the *RL-agent* using $\epsilon$-greedy policy. For every increasing time-step $t$, we assign the action $a_t$ from the maximum expected return at that time-step following policy $\pi$ that maps the state vector sequence $s_t$ with the permissible actions $a$ using *argmax* function. Following which we update the state of the RL in the next time-step $s_{t+1}$. For experience $e_t$ in memory set $D$, we update the parameters of the target action-value function $\hat{Q}(s, a)$. We use stochastic gradient descent (SGD) [25] optimization algorithm to update the network parameters of *RL-agent* $Q$ with the loss calculated as per Equation 7. In the Section 4.2 of the paper, we show the results of varying $\gamma$ and learning-rate in the RL-agorithm.

## 4 Experiments and Evaluation

In this section, we elaborate the data processing techniques implemented to carefully select the suitable input features for our SRU *predictor* and we also discuss the preliminary results of training the SRU with and without the attention mechanism.

### 4.1 Data Analysis and Preparation

In this section of the paper we describe the collection and pre-processing of the data from various vaccination drives in different states of the USA, and the simulation of cost matrix data considering various parameters in a cold chain distribution channel. This pre-processed data constitutes the total *state-space* of the *environment* of our RL Algorithm 2.

#### 4.1.1 Vaccination Drive Data Preparation for RNN based SRU model

The data[2] consists of state-wise Covid-19 vaccine administration details of USA, with date stamps from $1^{st}$ January to $9^{th}$ August, 2021. Figure 5 shows the daily number of doses of vaccines (*total_vaccinations*) administered per state. We observe the data distribution across 240 dates such that the vaccinations peak when the demand is highest in each state, and as the number of people vaccinated falls, the number of vaccinations administered falls. We find the population for each state from the official census of USA [3]. There are 240 sets of data points for each state. We analyse the correlation plots for all the given attributes collected from the public dataset in Figure 4. From this plot we observe that the feature attributes *total_vaccinations*, *total_distributed*, *people_vaccinated*, and *people_fully_vaccinated* (attribute numbers 0, 1, 2 and 5 in Figure 4) have the highest positive correlation with the target attribute *daily_vaccinations* (attribute number 8 in Figure 4). However, these feature attributes have high mutual positive correlation as well. Hence to prepare a better input feature space for our SRU *predictor* model we derive a new feature from the available feature attributes.

---

[2]https://ourworldindata.org/us-states-vaccinations
[3]https://www.census.gov

**Algorithm 2:** RL Algorithm

**Data:** $Cost_{state}$ sequence, $Demand_{state}$ sequence
**Result:** trained target action value function $\hat{Q}$ to give better rewards $r$

Initialize action-value function $Q$ and target action-value function $\hat{Q}$ with random weights $\vartheta$;
**for** $state \leftarrow 1$ *to* $S$ **do**
    Initialize $D_{state}$ to capacity N;
    $s \leftarrow ( Cost_{state}, Demand_{state} )$;
    **for** $episode \leftarrow 1$ *to* $M$ **do**
        $s_1 \leftarrow ( Cost_{state,1}, Demand_{state,1} )$;
        $\phi_1 \leftarrow \phi(s_1)$;
        **for** $t \leftarrow 1$ *to* $T$ **do**
            **if** $t = 1$ **then**
                Randomly select $a_t$ with a probability of $\epsilon$;
            **end**
            **else**
                $a_t \leftarrow argmax_a Q_t(s_t, a; \vartheta_1)$;
            **end**
            $s_{t+1} \leftarrow |s_t, a_t, |Cost_{state,t}, Demand_{state,t}|$;
            $\phi_{t+1} \leftarrow \phi(s_{t+1})$;
            $D \leftarrow |\phi_t, a_t, r_t, \phi_{t+1}|$;
            Random sample mini-batch $|\phi_k, a_k, r_k, \phi_{k+1}|$ with $N$ transitions from $D_{state}$;
            **for** $e_t = |s_t, a_t, r_t, s_{t+1}|$ **do**
                $\hat{Q}(s, a) = E[r + \gamma \max_a \cdot \hat{Q}(s', a')]$;
            **end**
            SGD (stochastic gradient descent) on $L_k(\vartheta_k)$ with respect to $\vartheta_k$ with pre-defined learning rate; Weights of $\hat{Q} \leftarrow$ weights of $Q$;
        **end**
    **end**
**end**

From the raw data attributes, *total_vaccinations* and *people_fully_vaccinated*, we calculate a new feature, *people_partially_vaccinated*. The *total_vaccinations* and *people_fully_vaccinated* are total number of Covid-19 vaccinations administered and total number of fully vaccinated people (first and second dose) respectively, in a particular state till a specific date. The formula to derive it is given by Equation 9.

$$N_S = total\_vaccinations - 2 \times N_D \quad (9)$$

In Equation 9, $N_S$ and $N_D$ represent the number of people vaccinated with a single dose and both the doses respectively. These two parameters along with the state population data ($X^1$, $X^2$, $X^3$ respectively) constitute the input to the *predictor* model. In Figure 6, we show the values of *Pearson's* correlation coefficients for the final input features to our SRU model. By ensuring a low correlation between the input variables, we create a stable dataset for the *predictor* model's training.

Hence, the input to the SRU model is given by the sequence of a set of 3 features, *total_population*, *people_partially_vaccinated*, and *people_fully_vaccinated*, from the first date to a selected vaccination date. The data is split into *80%*, *10%* and *10%* for *train-set*, *validation-set*, and *test-set* respectively. The ground truth for training and evaluating the SRU model is given by the *daily_vaccinations* target attribute values from the dataset.

### 4.1.2 Cost Matrix Data Preparation for State Space of Q-Learning

According to Thomas *et al.*[26], there are a number of major cost components in a health system temperature controlled supply chain network. These include the costs related to – 1. procurement, 2. storage, 3. transportation, and 4. management. This is a function-wise categorization of cost data. In tier-wise categorization of cost data, particularly in case of our analysis, we divide it into the costs related to – 1. Government Medical Store Depots (GMSDs), 2. state-wise and 3. regional costs. Generally in a vaccine distribution chain, vaccines from a centralized distribution location is supplied across various states in a country. These are referred to as GSMDs. In India, there are 7 GMSDs from which the vaccines are distributed all over the country. Since we only have state-wise vaccination data available, we only apply the functional categorization of supply-chain cost

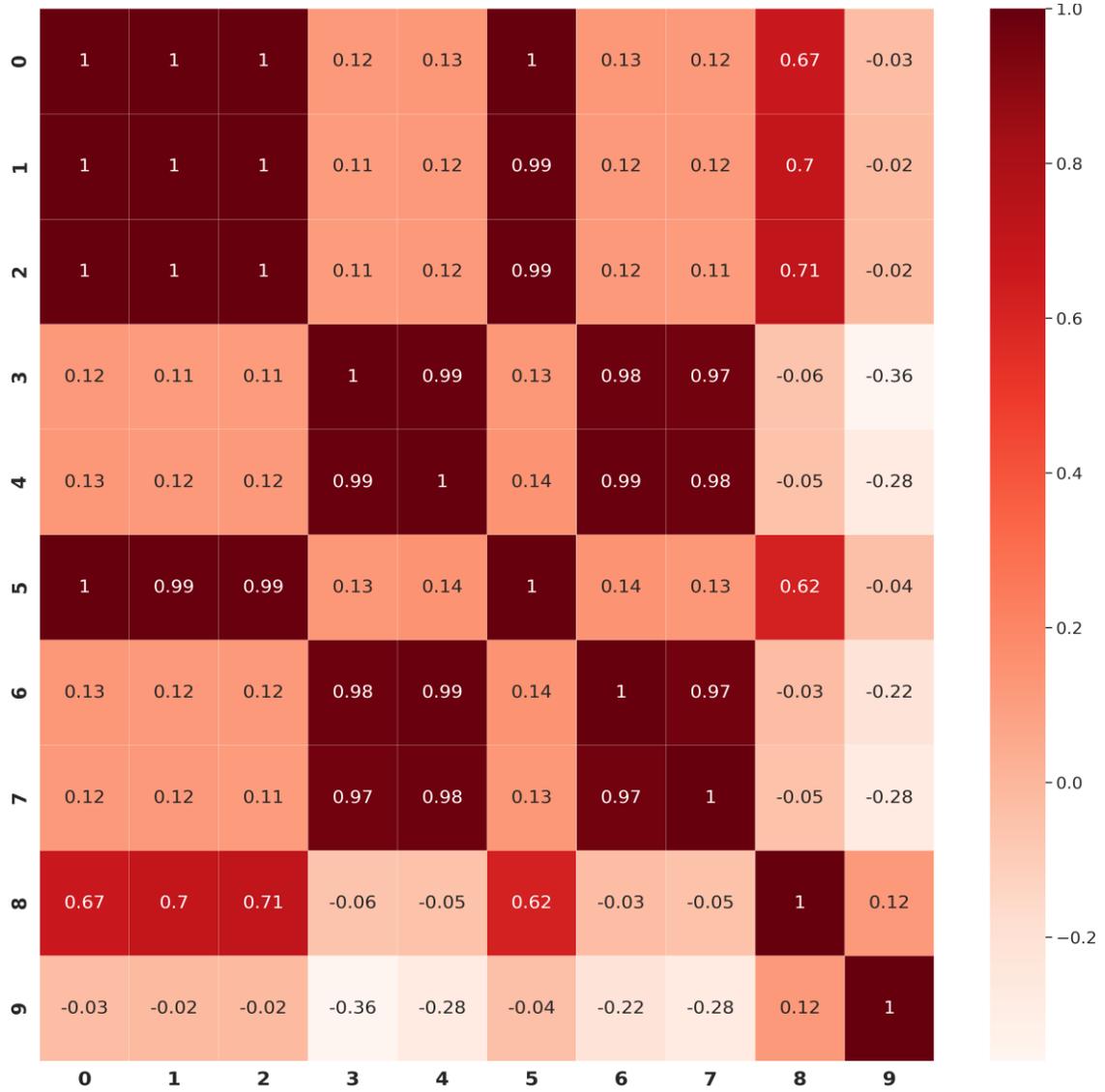

Figure 4: The mapping of the values in the x and y axes are as follows – 0 - total_vaccinations, 1 - total_distributed, 2 - people_vaccinated, 3 - people_fully_vaccinated_per_hundred, 4 - total_vaccinations_per_hundred, 5 - people_fully_vaccinated, 6 - people_vaccinated_per_hundred, 7 - distributed_per_hundred, 8 - daily_vaccinations and 9 - daily_vaccinations_per_million

components for state-wise tier category. To aid this explanation we provide a chart in Figure 8. Hence, for our cost data generation, we need the following major cost-components –

1. labor cost in procurement,
2. labor, space (inventory), equipment and fuel costs in storage,
3. labor, vehicles and fuel costs in transportation, and
4. labor and operational costs in management.

For each of the above parameters we simulate the relevant cost values for each state in the USA to create the cost-matrix $Cost_{state}$ in the *state-space* of our DQN *RL-agent*. This calculation of cost associated with vaccine supply in each state of the USA is done using equation 10. Here, we have assumed that the distribution to different states takes place from a common source location, which is generally the case in real world scenario.

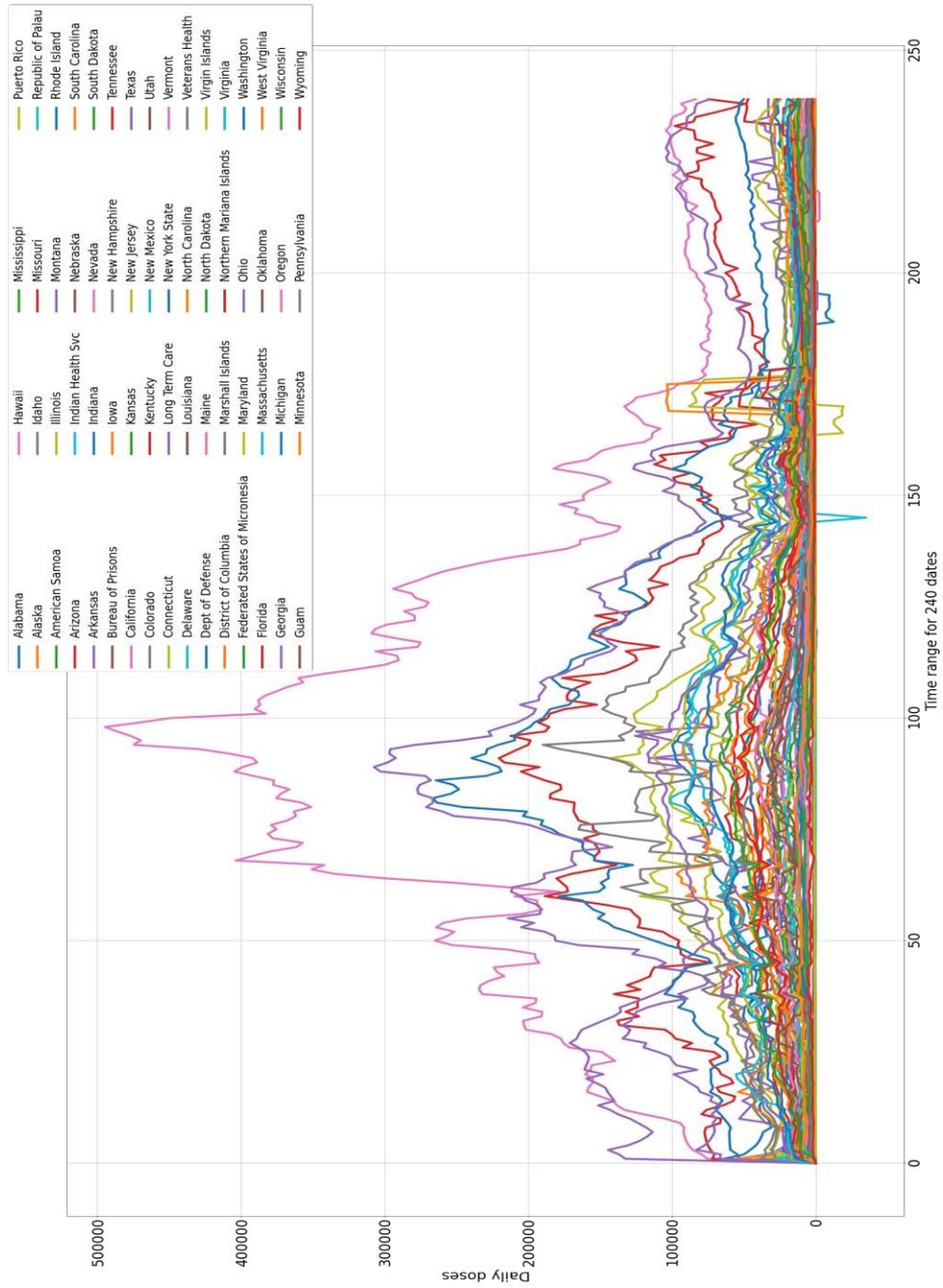

Figure 5: State-wise daily vaccinations vs time. The daily vaccinations gradually rise and fall indicating the gradual increase and decrease of demand as more people are vaccinated overtime.

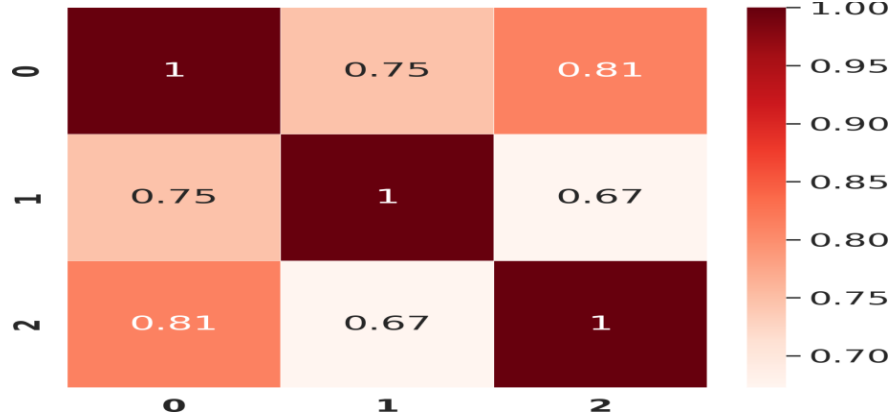

Figure 6: Pearson's correlation matrix plot for the set of inputs to the SRU model. The x and y axes values map as follows – 0 - total_population, 1 - people_partially_vaccinated and 2 - people_fully_-vaccinated.

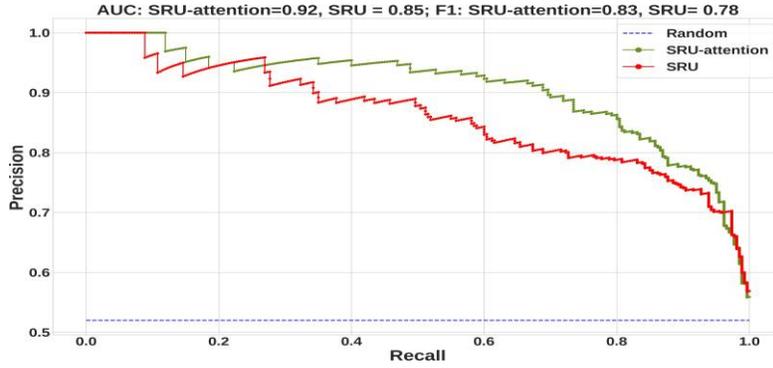

Figure 7: Precision-Recall (PR) curve of VacciNet's *predictor* SRU, with and without attention.

$$Cost_{state} = \sum_{i=proc,stor,trans,mang} (\Pi_{j=labor,fuel,space,equip,vehicle}(W_{i,j} * Cost_{i,j}) / \sum_{i} W) \quad (10)$$

In equation 10, *proc*, *stor*, *trans*, *mang* refers to the costs associated with procurement, storage, transportation and management respectively. We simulate the cost matrix $Cost_{i,j}$ by generating the values for each state. This simulated data takes into account the distances of the states from the centre of distribution, the general financial condition of the state, the unemployment ratio, and the budget allocation of the state towards Covid-19 pandemic. Although this data is simulated to replicate a real world scenario, the supply-chain prices highly vary in the real case. However, with the availability of the cost of distribution data, our framework can be re-trained on it to provide good results. To ensure further variations in the final cost per state, we introduce a weight element for each cost element. Here, $W_{i,j}$ represents the total weight matrix with few of the weight elements assigned zero value purposefully. For example, when there is no cost associated with space or inventory in procurement, the relevant weight element will be zero in value. The other values of weight elements are randomly simulated ranging from $> 0$ to $1$. This helps us calculate a weighted average of the cost values which is normalized to get a normal distribution of numbers. This is an additional pre-processing step for better training of the DQN *RL-agent*. This normalized weighted average cost values along with the vaccine demand predictions from the SRU model form the environment state space of the RL framework.

### 4.2 Model Evaluation

In this section of the paper, we provide the details of the evaluation of our two learning methods - supervised learning of RNN-based SRU *predictor* model and reinforcement learning of DQN *RL-agent*.

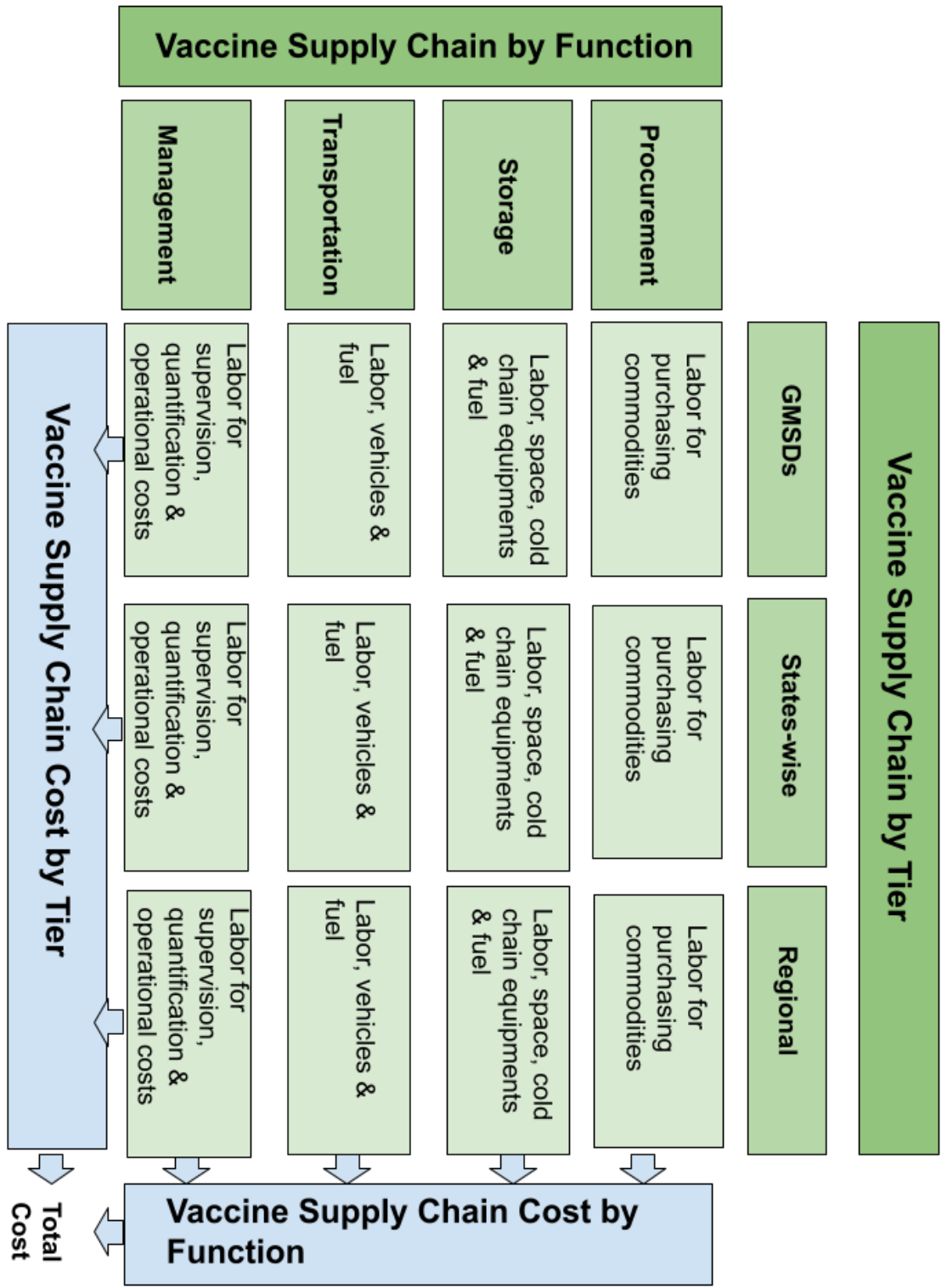

Figure 8: Supply chain cost categorization for cold chain distribution of vaccines.

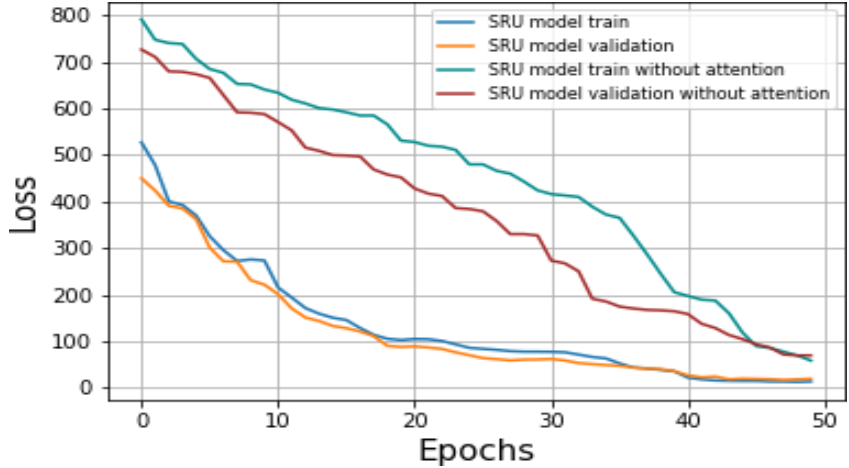

Figure 9: The SRU without the attention network starts at a higher loss. The SRU with the attention mechanism converges 2.9x faster than SRU without attention network.

#### 4.2.1 SRU *predictor* evaluation

The SRU *predictor* does not require more than 10 dimensionally expanded features to learn better representations. Hence, for simplicity of the architecture we limit the number of SRU units to 10 per layer. The attention mechanism aids in constructing fixed size context vectors from the encoded representations. The context vector improves the decoding step performed by the single dense/linear layer. Without the attention mechanism, the model takes a much longer time to converge to lower values as evident from Figure 9.

Without the attention mechanism, the input of the sequences is required to be fixed to 240x3 (240 sequences of 3 features each) so that the input to the dense layer would always be of size $240 \times 10 = 2400$ (10 feature outputs per SRU layer). The sequences are required to be padded as well resulting in excess of zeros that contributes to the poor learning, which is solved using attention. It also enables variable length inputs. As shown in Figure 7, Precision-Recall (PR) curve of SRU *predictor* model with attention mechanism has better Area Under Curve (AUC) and F1-score (F1) than that without attention on test dataset. With *attention* block, the AUC increase from $0.85$ to $0.92$ and F1-score increase from $0.78$ to $0.83$. For time-series data, with a continuous target variable it is a challenging task to reach good accuracy. We determine the accuracy in terms of F1-score with a pre-defined threshold to determine how close the predicted values are to the ground truth. We defined the allowable deviation of predicted output in the range of $10\% - 20\%$ with respect to the target variable.

### 4.3 DQN *RL-agent* evaluation

We train our RL framework based on Algorithm 2 for each of different states in USA. In our experiments we initialize the exploration probability to $1$ and reach minimum exploration probability value of $0.001$ with an exponential decay rate of $0.0001$. We train our deep Q-network on $10000$ episodes with $1000$ iterations in each. We use different initial learning rates as shown in Figure 10, and in each case observe the reward for different discount factors ranging between $0.1$ and $0.9$. From the figure we observe that in all the cases, discount factor 0.9 is giving the highest rewards. We observe the min-max scaled average reward per state transition is $0.678$ with initial learning rate $0.1$ and $\gamma$ value $0.9$. This is the best performing model we inferred from our problem formulation.

## 5 Discussion and Conclusion

There exists a number of earlier works based on deep learning methods for optimization of supply chain management tasks, especially in the cold chain networks which require refrigeration and hence contribute immensely in the total supply cost. Some of the earlier works in this area include implementation of block chain technology [27] [28] [29]. Nikolopoulos *et al.*[30] implemented time series models, regression models, LSTMs (Long-Short Term Memory) [19] to forecast supply demands. LSTMs are popularly used in several earlier works for supply chain management solutions [31] [30] [32] [33]. Although LSTMs are the state of the art deep learning models in time-series data analysis, they are difficult to train (large training time) and deploy. Reinforcement learning (RL) techniques are being used in supply chain network optimization works for a long period of time as well [34] [35] [36]. Among several RL techniques, Q-learning has however over the years proven to be more efficient in terms of training complexity and deploy-able model performance metrics. Traditional Q-Learning with Q-Tables have been replaced by Deep Q-Learning with deep neural networks (Multilayer Perceptrons or MLPs) [37] called deep Q-Networks (DQNs) [9]. Deep Q-learning has risen to be more accurate in predictions given sufficient data. DQNs are a more energy and memory efficient implementations of RL. Deep Q-Networks have been used for demand-supply synchronization [38]. Although there exists some studies on supply chain optimization techniques, the scenarios change in pandemic situation. There is a gap in research related to analysing the cost of essential item supply according to the demand during socio-economic crisis. It is critical to reach the right trade-off between cost of supply and demand of supply. Due to the recent pandemic in the year of 2020, we took into notice an otherwise overlooked flaw in the world-wide healthcare infrastructure. During the sudden upsurge of massive demands for vaccines and medicines across various nations, the world still relied on age-old techniques to strategize vaccine allocations. After the priority vaccination procedure was completed, the mass vaccination drives brought in new challenges where human errors in the supervision of the vaccine distribution networks caused major financial damages to countries. Every nation allotted huge budget towards vaccination drives, yet there was supply shortages in some vaccination centres while others faced large storage costs, or reshipment costs. Furthermore, there was uncertainty and indecision related to partial and full vaccinations. To the best of our knowledge, our work is the first of a kind that addresses all these issues. We propose a framework that is capable of reaching the right trade-off between cost and demand. In the future scope of our work, the VacciNet framework can be scaled to optimize a denser vaccine distribution chain provided more granular data of exact locations of medical centers, pharmacies, vaccination centres, etc. Our RL based learning technique gets better with the amount of data it trains on. There is a lack of publicly available data on vaccination, vaccine supply and supply cost. Provided data from many more countries, and more granular data on supplies to cities and local pharmacies, our framework can be further trained for robustness.

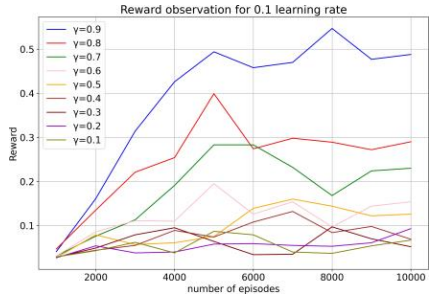 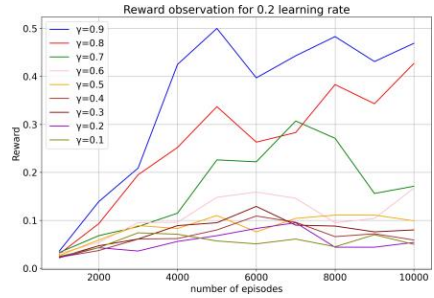

(a) For learning rate = 0.1

(b) For learning rate = 0.2

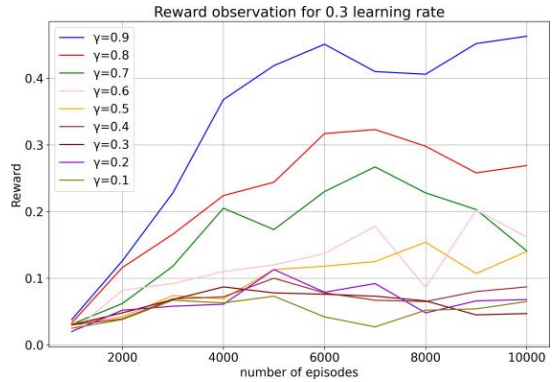

(c) For learning rate = 0.3

Figure 10: The reward observations with respect to progressing episodes of deep Q-Learning with varying $\gamma$ values (discount) for different learning rates.